\documentclass[review]{elsarticle}
\usepackage{lineno,hyperref}
\usepackage{graphicx}
\modulolinenumbers[5]
\usepackage{array}
\usepackage{wrapfig}
\usepackage{multirow}
\usepackage{tabularx}
\usepackage[utf8]{inputenc}
\usepackage{booktabs}
\journal{Information Processing and Management}
\usepackage{multirow}
\usepackage{graphicx}
\usepackage{url}
\usepackage{color,soul}
\pdfoutput=1

\UseRawInputEncoding

\biboptions{authoryear}

\begin{document}

\begin{frontmatter}

\title{Detecting White Supremacist Hate Speech using Domain Specific Word Embedding with Deep Learning and BERT}
\author{Hind Saleh, Areej Alhothali, Kawthar Moria
\\
\normalsize{Department of Computer Science, King Abdulaziz University,KSA},
\normalsize{ halatwi0003@stu.kau.edu.sa.,aalhothali,kmoria@kau.edu.sa}
}
\begin{abstract}
 White supremacists embrace a radical ideology that considers white people superior to people of other races. The critical influence of these groups is no longer limited to social media; they also have a significant effect on society in many ways by promoting racial hatred and violence. White supremacist hate speech is one of the most recently observed harmful content on social media. Traditional channels of reporting hate speech have proved inadequate due to the tremendous explosion of information, and therefore, it is necessary to find an automatic way to detect such speech in a timely manner. This research investigates the viability of automatically detecting white supremacist hate speech on Twitter by using deep learning and natural language processing techniques. Through our experiments, we used two approaches, the first approach is by using domain-specific embeddings which are extracted from white supremacist corpus in order to catch the meaning of this white supremacist slang with bidirectional Long Short-Term Memory (LSTM) deep learning model, this approach reached a 0.74890 F1-score. The second approach is by using the one of the most recent language model which is BERT, BERT model provides the state of the art of most NLP tasks. It reached to a 0.79605 F1-score. Both approaches are tested on a balanced dataset given that our experiments were based on textual data only. The dataset was combined from dataset created from Twitter and a Stormfront dataset compiled from that white supremacist forum.  
\end{abstract}
\begin{keyword}
\texttt{white supremacist}\sep  NLP\sep deep learning \sep BERT \sep classification
\end{keyword}
\end{frontmatter}
\section{Introduction}
\noindent
Social media has become an essential element of our society by which people communicate and exchange information on a daily basis. The strong influence of social media on internet users has been of great benefit to many individuals, businesses, and organizations. Many companies and organizations nowadays use social media to reach customers, promote products, and ensure customer satisfaction. Despite the benefits associated with the widespread use of social media, they remain vulnerable to ill-intentioned activities, as the openness, anonymity, and informal structure of these platforms have contributed to the spread of harmful and violent content.
\par 
Although social media service providers have policies to control these ill-intentioned behaviors, these rules are rarely followed by users. Social media providers also allow their users to report any inappropriate content, but unreported content may not be discovered due to the huge volume of data on these platforms. Some countries have restricted the use of social media, and others have taken legal action regarding violent or harmful content that might target particular individuals or communities. However, these violations might end up unpunished due to the anonymous nature of these platforms, allowing ill-intentioned users to fearlessly share harmful content by using nicknames or fake identities. One of the most-shared harmful content on social media is hate content, which might take different forms such as text, photos, and/or video. Hate speech is any expression that encourages, promotes, or justifies violence, hatred, or discrimination against a person or group of individuals based on characteristics such as color, gender, race, sexual orientation, nationality, religion, or other attributes~\cite{weber2009manual}. Online hate speech is rapidly increasing over the entire world, as nearly $60$\% of the world’s population ($\approx$ $3.8$ billion) communicates on social media~\cite{WeAreSocial}. Studies have shown that nearly $53$\% of Americans have experienced online hate and harassment~\cite{AntiDefamationLeague}. This result is $12$\% higher than the results of a comparable questionnaire conducted in $2017$ ~\cite{duggan2017online}. For younger people, the results show that $21$\% of teenagers frequently encounter hate speech on social media~\cite{JClement}. 
\par 
 One of the most dangerous and influential forms of online hate speech is led and spread by supporters of extreme ideologies who target other racial groups or minorities. White supremacists are one of the ideological groups who believe that people of the white race are superior and should be dominant over people of other races; this is also referred to as white nationalism in more radical ideologies~\cite{blazak2009toward}. White supremacists claim that they are undermined by dark skin people, Jews, and multicultural Muslims, and they want to restore white people’s power, violently if necessary. They have also claimed responsibility for many violent incidents that happened in the $1980$s, including bank robberies, bombings, and murders~\cite{blazak2009toward}. The white supremacist ideology has been adopted by both right-wing and left-wing extremists who combine white supremacy with political movements~\cite{ivan2019extremist,DrewMillard}.
\par 
 White supremacist hate speech has become a significant threat to the community, either by influencing young people with hateful ideas or by creating movements to implement their goals in the real world. A study has also suggested links between hate speech and hate crimes against others (e.g., refugees, blacks, Muslims, or other minorities)~\cite{williams2020hate}. Several recent brutal attacks have also been committed by supporters of radical white supremacists who were very active members on social media. The mass shootings in New Zealand~\cite{cai_landon_2019}, Texas~\cite{cai_landon_2019}, and Norway~\cite{BBCNews} were committed by white supremacists who had shared their opinions and ideologies on social media. The attacker of two mosques in Christchurch, New Zealand, was a 28 year old man who identified himself as a white nationalist hero~\cite{cai_landon_2019}, and posted a manifesto that discussed his intent to kill people as a way to reinforce the sovereignty of white extremists. From a psychological point of view, any violent attack must be preceded by warning behaviors, which includes any behavior that shows before a violent attack that is associated with it, and can in certain situations predict it. Warning behaviors can be either real-world markers (e.g., buying a weapon and make a bomb) or linguistic markers or signs (e.g., “I had a lot of killing to do”) which can happen in real life and/or online~\cite{cohen2014detecting}. 
\par 
 Automatic detection of white supremacist content on social media can be used to predict hate crimes and violent events. Perpetrators can be caught before attacks happen by examining online posts that give strong indications of an intent to make an attack. Predicting violent attacks based on monitoring online behavior would be helpful in crime prevention, and detecting hateful speech on social media will also help to reduce hatred and incivility among social media users, especially younger generations.
\par 
Studies have investigated the detection of different kinds of hate speech such as detecting cyberbullying ~\cite{dadvar2013improving,huang2014cyber,haidar2016cyberbullying}, offensive language ~\cite{chen2012detecting,pitsilis2018detecting}
, or targeted hate speech in general by distinguishing between types of hate speech and neutral expressions~\cite{ribeiro2017like,ribeiro2018characterizing,djuric2015hate}. Others have dealt with the problem by detecting a specific types of hate speech, such as anti-religion~\cite{albadi2018they,zhang2018detecting}, jihadist~\cite{de2018automatic,ferrara2016predicting,wei2016identification,gialampoukidis2017detection}, sexist, and racist~\cite{pitsilis2018effective,badjatiya2017deep,gamback2017using}. However, less attention has been given to detecting white supremacism in particular, with limited studies~\cite{de2018hate}.  
\par
\begin{figure}[htbp]
\centerline{\frame{\includegraphics[scale=.5]{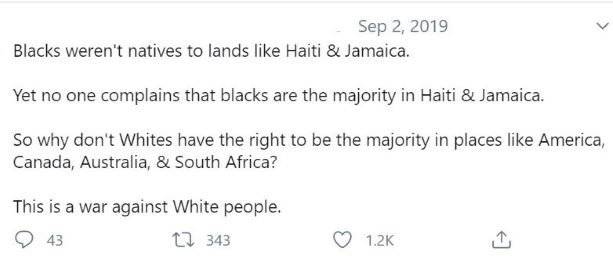}}}
\caption{Example of white supremacist when they claim they are undermined.}
\label{fig:fig2}
\end{figure}

 
White supremacist extremists tend to use rhetoric (i.e., the art of effective and compositional techniques for writing and speaking) \cite{brindle2016language} in their language. They also use specific vocabulary, abbreviations, and coded words to express their beliefs and intent to promote hatred or encourage violence to avoid being detected by traditional detection methods. They mostly use hate speech against other races and religions, or claim that other races are undermining them. Figure~\ref{fig:fig2} shows an example of a white supremacist tweet.
\par 
\subsection{Research goal and contributions} 
In this paper, we aim to detect white supremacist tweets based on textual features by using deep learning techniques. We collected about $1M$ tweets from white supremacist accounts and hashtags to extract word embeddings, and then we labeled about $2k$ subsets of the data corpus to build a white supremacist dataset. We applied two approaches: the first uses domain-specific word embedding learned from the corpus and then classifies  tweets using a Bidirectional LSTM-based deep model. This approach is evaluated on multiple dataset and achieved different results depending on the datasets that ranged from a $49.2$\% to a $74.8$\% F1-score. The second approach uses a pre-trained language model that is fine-tune on the white supremacist dataset using Neural Network dense layer. The BERT language model F1-scores ranged from $58.7$\% to $79.6$\%.
Thus, the research contribution can be summarized as follow:
\begin{enumerate}
\item Assessing the performance of domain-specific embeddings with bidirectional LSTM based deep model. 
\item Providing to the community experiments and results of domain-specific embeddings with deep models on white supremacist detection.
\item Providing a small dataset of English tweets with most recent white supremacist speech.
\item Performing experiments using BERT models for white supremacist detection, accordingly, providing important baselines for future work comparison.
\end{enumerate}

\par 
The rest of the paper proceeds with the Background Section (Section~\ref{sec:Back}), which provides information on the methodology used, related studies in the Literature Review section (Section~\ref{Sec:LR}), a detailed description of methods in the Methodology section (Section~\ref{sec:Meth}), details of the used datasets in the Dataset section (Section~\ref{sec:DS}), specifications of the methodologies and the results of each approach in the Experiments and Results section (Section~\ref{sec:ER}), observations and analysis of the performance of each approach in the Discussion section (Section~\ref{sec:DIS}), and finally, the Conclusion and Future Work section (Section~\ref{sec:CON}).  

\section{Background}
\label{sec:Back}
\noindent 
This section provides background information on the state-of-the-art methodologies used for natural language processing (NLP) tasks, includes the current commonly used pre-trained embedding (i.e., word representation) and language models. For pre-trained word embeddings, different organizations and institutions (e.g., Google, Stanford) continuously seek to find the best methods for word representations (word meaning). Here, we describe the most commonly used word embedding (word representation) model according to recent studies. Pre-trained language models have recently received massive attention in the  NLP field. They can be defined as a black box that understands natural language and can be applied and fine-tuned to solve NLP tasks. The pre-training process uses inexpensive unlabeled data to learn the initial parameters of a neural network model. Bidirectional Encoder Representations from Transformers (BERT) is one of these language models and is state of the art for many NLP problems.

\subsection {Pre-trained Word Embedding}
\label{Subsec:Pretrain}
\noindent 
Word embedding~\cite{bengio2003neural} is one of the most popular recent Natural Language Processing (NLP) trends. It refers to any technique aiming to map words to a dense vector representation that captures words semantic meanings that can be used to estimate similarities between words for classification~\cite{liu2018neural}. The primary purpose of this mapping is to represent linguistic terms in dense vectors to be utilized by machine learning algorithms. A word is mapped to an N-dimensional vector appropriate for representing the meaning of a specific language. Different Neural Network (NN) models have been used to construct word vectors~\cite{mikolov2013efficient}, as word vectors provide meaningful numerical descriptions of words based on their context~\cite{liu2018neural}.

\par 
Word embedding has proven to be a powerful technique for extracting the most meaningful representations of words. The evolution of word embedding has resulted in tremendous success in various NLP tasks like text classification~\cite{gamback2017using,lilleberg2015support}, document clustering~\cite{ailem2017non}, part of speech tagging~\cite{wang2015part}, named entity recognition~\cite{sienvcnik2015adapting}, sentiment analysis~\cite{tang2014learning,wang2016dimensional,al2017using}, and so on. Many researchers have built models to reach the best meaningful word vector representations by using word embedding, and the most common models are Google Word2Vec and Stanford GloVe.

\subsubsection {Word2Vec}
\label{Subsec:w2v}
\noindent 
Word2Vec developed by Google research team~\cite{mikolov2013efficient} to overcome traditional word representation (vector space) techniques by representing words in a more dense and meaningful representation given a corpus context. The word vector representation is computed from a large corpus fed into the model to produce vectors representing word meanings. The meaning of words is obtained from surrounding words within a specified window size. Word2Vec representation can be obtained from different model architectures, e.g., continuous skip-gram and continuous bag-of-words (CBOW). Google released a pre-trained Word2Vec model representing word meanings that have been successfully utilized in many NLP tasks in recent years. The model is trained on a vast corpus of $100$ billion words and is publicly available~\cite{Googleword2vec}. A word vector is more meaningful if the model is trained on a progressively larger corpus size.
\subsubsection {GloVe}
\label{Subsec:glv}
\noindent 
GloVe (Global Vectors for Word Representation) is another word embedding model developed by~\cite{pennington2014glove}. It is an unsupervised learning algorithm that obtains a vector representing a word’s semantic meaning by using corpus-based distributional features. The algorithm performs several operations on a constructed word-to-word co-occurrence statistics-based matrix. This is a costly process for a huge corpus, even though it only requires a single pass through the corpus. This matrix is used to construct word vectors instead of using a prediction-based approach like in Google Word2Vec. Thus, the main difference between Google’s Word2Vec and GloVe is that Word2Vec is a prediction-based model in which a loss function is used to evaluate the prediction performance, while GloVe is a count-based model. GloVe has been trained on many platforms such as Wikipedia, web crawl data, and Twitter, and provides a model for each one with different dimensions~\cite{pennington2014glove}.

\subsection{BERT pre-trained language model}
\label{Subsec:BERT}
\noindent 
Bidirectional Encoder Representations from Transformers (BERT)~\cite{devlin2018bert}, is the latest revolution in NLP pre-trained language model trends. BERT is a deeply bidirectional language model trained on very large datasets (i.e., Books corpus and Wikipedia) based on contextual representations. Other previous language models are unidirectional, which means they consider context only from left-to-right or right-to-left, whereas BERT adds a Neural Network Dense layer for classification to construct a fully pre-trained language model ready for fine-tuning. The fine-tuning advantage incorporates the contextual or the problem-specific meaning with the pre-trained generic meaning and trains it for a specific classification problem. BERT provides high performance for NLP tasks and improves on the results from traditional models. 
\par 
NLP tasks seek to find the best contextual word representations. Word2Vec and GloVe generate an embedding representation for each token, regardless of its contextual differences, and a word's meaning is changed according to its associated context. If the word has different meanings based on context, GloVe and Word2Vec represent the word as a single embedding, such as the word 'bank' in the phrases bank deposit and river bank. Here, the word bank would have a single representation in the whole corpus, which ignores other meanings of the word. Therefore, Word2Vec and GloVe are described as context-free models. Contextual representation has two types: unidirectional, in which the representation is learned in one direction, from left to right, or bidirectional, in which the representation of the word in learned from both directions, i.e., left to right and right to left. BERT is deeply bidirectional by jointly conditioning both left and right contexts in all layers~\cite{devlin2018open}. BERT models have different releases that differ according to model size, cased or uncased alphabet, languages, and the number of layers, and they are all available online.
\subsection{Deep Learning}
\label{Subsec:DL}
\noindent
Deep learning (also known as layered representations learning and hierarchical representations learning) is a subfield of machine learning which uses successive layers for accurate representations of meaning~\cite{chollet2018deep}. The learning process is performed by exposing training data to the model to give representations. If the learning model consists of only one or two layers, then it is called shallow learning. Deep learning usually uses a neural network in order to learn these representations, and the neural networks are structured in layers. The learning process aims to find the best-weight values of the neural network that map an input example to its correct target, and a loss function is used, which measures the distance between the predicted and actual targets. Different constructions of layers give different deep learning models. Neural networks form the basis for deep learning, and one of the most common neural network architectures used for deep learning construction is LSTM.
\subsubsection{Long Short-Term Memory (LSTM)}
\label{Subsec:LSTM}
\noindent
LSTM is a recurrent neural network (RNN) developed as a solution for solving the problem of vanishing gradient in RNNs~\cite{hochreiter1997long}. An RNN is a specific type of neural network which considers the history or context in the computation of the output. RNN includes a memory to preserve the previous computational result and feeds back the previous set of hidden unit activation to the network with the current input. This particular architecture is beneficial for problems that require the history to be involved in the decision-making process, such as speech recognition and stock forecasting. RNNs suffer from the vanishing gradient problem, in which the weights are lost in a deeper layer of the network, thereby failing to capture very long dependencies. To avoid this problem, LSTM replaces each node by a memory cell, which consists of an input gate, forget gate, output gate, and a node connected back to itself. The memory cell in a specific layer uses the hidden state in the previous layer during the current time and the hidden state of the current layer from the previous time. The forget gate decides which information should be ignored in the cell state, and the input gate and tanh layer decide which information is stored in the cell state, then using the sigmoid function to decide the final output~\cite{liu2018neural}. 

\section{Literature Review}
\label{Sec:LR}
\noindent This literature review covers prominent studies related to hate speech detection. The widespread use of social media by people worldwide has contributed to the increase in hate speech and other problems that the current research seeks to solve. There has been a considerable research effort with regard to hate speech detection, but not much effort into specifically detecting white supremacist hate speech.  
\par
 ~\citet{liu2018neural} introduced hate speech word embedding to achieve higher accuracy in hate speech detection, and achieved $78$\% by using word embeddings trained on Daily Stormer articles and high centrality users' tweets. It was concluded that a conventional neural network (CNN) performed better than LSTM on tweets because of the shorter sentences; however, the study was based on tweets $140$ characters long, but Twitter then extended tweet lengths to $280$ characters. Word embeddings provide an alternative solution for the traditional features in hate speech detection. A comparative study was conducted~\cite{gupta2017comparative} to assess the performance of the Word2Vec model to detect hate speech on three datasets, and achieved maximum accuracy of a $0.912$ F1-score given that they used average Word2Vec embeddings and an logistic regression classifier. They concluded that domain-specific word embedding provides better classification results and is suitable for unbalanced classes.  
 ~\citet{nobata2016abusive} used pre-trained word embeddings and a regression model to detect abusive language, and achieved a $60.2$ F1-score on a finance domain and a $64.9$ F1-score on a news domain, but Word2Vec provided better performance, with $5$\% on both domains. ~\citet{badjatiya2017deep} used deep learning from domain-specific learning tuned towards hate speech labels to extract features, and used a decision tree as a classifier. Their best F1-score was $93$\%, and by using random embeddings and an LSTM combination for features, they reported that domain specific embeddings learned using deep neural networks expose “racist” or “sexist” biases for various words. From the above studies, domain- specific-based detection has good performance due to it providing more domain-related hate words frequently used by users in a given domain. 
 
\par 
Several studies have looked into detecting online hate speech that is similar to white supremacy in targeting others based on their racial and cultural identities. ~\citet{hartung2017identifying} classified Twitter profiles into either right-wing extremist or not. They used a bag of words for lexical features, emotional features, lexico-syntactic, and social identity features, and SVM with a linear kernel to classify them. The reported an F1-score of $95$\%, which was achieved by using a bag of words feature which outperformed all other features combined. They reported the most common features for each class by performing a qualitative analysis in the German language (e.g., asylum seekers, citizens' initiative, demonstration, and autumn offensive). They found that the content of tweets is a good indicator for hateful accounts. However, the study was limited to the German language, and they used traditional features (e.g., bag-of-words) with machine learning classifiers.
\par  ~\citet{hartung2017ranking} aimed to identify German right extremist accounts, and the main task is to rank unknown profiles based on their relative proximity to other users in the vector space. They used four feature sets: lexical (word stems), social identity, emotional, and lexico-syntactic (sentence constructions). The proposed model represented each Twitter profile as a point in a high-dimensional vector space based on the feature set. The result of the classification was a $65$\% f1-score obtained by using an unbalanced discrete decoding model over all the subsamples. The results also showed that the f1-score increased to $81$\% when the profiles had greater than $100$ tweets. This shows that the proposed ranking model depends heavily on the number of tweets of the profile. However, this condition may not apply to extremist profiles as they often use newly created accounts, as found in other studies~\cite{ribeiro2017like}. Thus, the ability to detect extremist accounts is not valuable enough because of the number of tweet constraints (e.g., at least $100$ tweets), as this is a known characteristic of extremist accounts.
\par
The most recent and related study focusing on detecting white supremacist content in the Stormfront forum was done by ~\cite{de2018hate}. Their model was trained and tested on a balanced subset of a main dataset consisting of about $2k$ sentences collected from the Stormfront forum. Several machine learning approaches were examined, including SVM~\citep{hearst1998support}, Conventional Neural Network (CNN)~\citep{kim2014convolutional}, and LSTM~\citep{hochreiter1997long}, and the results showed that the LSTM outperforms the other models with an accuracy of $78$\% and $73$\% with and without excluding sentences that require extra context, respectively. The main limitation in this study is annotating sentences extracted from paragraphs without providing any additional knowledge that might help to understand the context of the sentences to label them accurately. The study reported the terms most used by users (e.g., ape, scum, savage), and also purported the experiment as a baseline for further investigations in the future.  
\section{Methodology}
\label{sec:Meth}
\noindent
We used two methods to investigate the detection of white supremacist hate speech: domain-specific word embedding with a deep model (Bidirectional LSTM), and BERT-based detection. Domain-specific word embedding is able to detect most terms, abbreviations, and intentional misspellings related to the white supremacist hate community which are not detectable by the general embedding model, since it is trained on books and Wikipedia textual data, which are often without misspellings. However, we also used BERT because it has proved to provide the state of the art for most current-day problems and even for some domain-specific problems~\citep{devlin2018bert}.
\subsection{Domain-Specific White Supremacy Word Embedding and Deep Model}
\noindent
To apply domain-specific embedding, we implement a sequence of steps, and for each step we specify the used methodology, as follows:
\subsubsection{Data Collection and Analysis}
\noindent Domain-specific word embedding involves word representations constructed from a corpus of a specific domain (e.g., politics, finance, sports). As mentioned earlier, using white supremacist domains to extract embedding helps to identify words that are commonly used in their community.  
\par
\begin{figure}[htbp]
\centerline{\includegraphics[scale=.7]{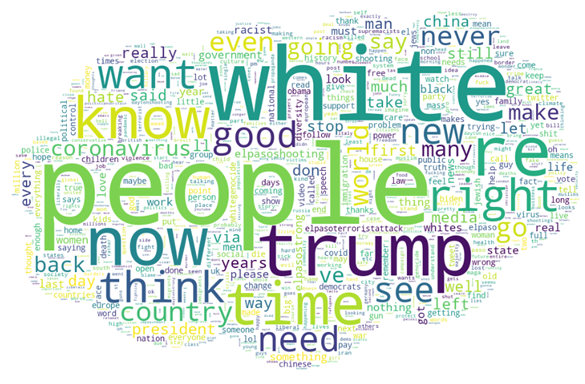}}
\caption{Word cloud of most used terms of white supremacist corpus.}
\label{fig3}
\end{figure}

To create the domain-specific word embedding, we first collected a corpus consisting of 1,041,576 tweets. The tweets were obtained from known white supremacist hashtags such as \#white\_privilege, and \#it\_is\_ok\_to\_be\_white. We also collected from accounts that explicitly (e.g., @Whit***er) or implicitly (e.g. @Na***st) identified themselves as white supremacist and\/or shared supportive phrases for white supremacy in hashtags or tweets encouraging or promoting racial or religious hatred against others. The White Supremacist Corpus (WSC) does not necessary include only white supremacist hate speech, and may include everyday tweets such as “good morning”. Then, we analyzed the corpus data to have an overall look at the most-used terms in that corpus. Figure~\ref{fig3} shows the terms most commonly used by their community, and they are different from hate speech terms.
\par 
We also analyzed the influence of using domain-specific word embedding of white supremacist hate speech by using word similarity. Word similarity measures the distance between the desired word vector and the closest other word vectors. Table~\ref{table:t1} includes the data analysis of our domain specific pre-trained model and general Google Word2Vec, GloVe models. The results show significant differences between the word embedding models. As can be seen, the words appearing in the domain-specific model tend to be more racist, while the Word2Vec and GloVe models provide the general meaning of a  word; for example, “Black” and similar words in domain-agnostic models tend to simply refer to the color of an object, while with domain-specific embedding the word “African” appears, which tends to be used mostly in hate speech. This observation confirms ~\cite{badjatiya2017deep} results.  

\begin{table}[]
\caption{Word similarity of word embedding models}
\label{table:t1}
\resizebox{\textwidth}{!}{%
\begin{tabular}{|l|l|l|l|}
\hline
\textbf{The word} & \textbf{Google Word2Vec} & \textbf{GloVe} & \textbf{Domain specific} \\ \hline
Black &
  \begin{tabular}[c]{@{}l@{}}white, Responded\_Letterman\_How, \\ blacks, crypt\_inscribed, transporting\_\\ petrochemicals, brown\end{tabular} &
  \begin{tabular}[c]{@{}l@{}}white, dark, blue, brown, red, \\ colored\end{tabular} &
  \begin{tabular}[c]{@{}l@{}}blacks, white, asian, Hispanic, \\ latino, latinx, african\end{tabular} \\ \hline
Muslim &
  \begin{tabular}[c]{@{}l@{}}muslims, Muslim, Moslem, islamic,\\  moslem, christian\end{tabular} &
  \begin{tabular}[c]{@{}l@{}}muslims, moslem, Islamic, \\ sunni, shiite, moslems\end{tabular} &
  \begin{tabular}[c]{@{}l@{}}Muslims, Islamic, somali, \\ pakistani, hindu, islamist\end{tabular} \\ \hline
Race &
  \begin{tabular}[c]{@{}l@{}}races, Race, racing, sprint,\\  Rain\_postpones\_Martinsville, Races\end{tabular} &
  \begin{tabular}[c]{@{}l@{}}races, racing, winner, finish, \\ event, runner\end{tabular} &
  \begin{tabular}[c]{@{}l@{}}races, ethnicity, racial, existence,\\  ethnicities, tribalism\end{tabular} \\ \hline
\end{tabular}%
}
\end{table}

\subsubsection{Pre-processing} 
\noindent 
Since we are dealing with real-world textual data, pre-processing techniques must be used to remove noise and exclude unrelated words. In this study, we only used standard pre-processing techniques without handling spelling mistakes or stemming. We performed the standard noise removal only since some words (e.g.\@, fck\@, f**k) are mostly intentionally misspelled or abbreviated to avoid being detected. 

\subsubsection{Feature Extraction (Word2Vec)}
\noindent
To build our domain-specific embedding from the white supremacy textual data, we trained the Word2Vec model on the collected white supremacy tweets. The training was performed using the Gensim library with the Continuous Bag of Word (CBOW) model, a window size of five words, and a 300-vector size. The CBOW model aims to predict a target word from its neighboring words. The result of this stage is word embeddings of the corpus words (i.e., the vocabulary). The domain-specific embedding method in this study will be referred to as White Supremacy Word2Vec (WSW2V).    
\subsubsection{Deep Learning Model}  
\noindent 
An extensive number of experiments to construct the most suitable deep learning model have been done by either changing the structure, the depth, or the parameters of models by using GridSearch. Based on these experiments, and to the best of our knowledge, the type of the deep model is sequential, and the most suitable structure consists of four layers. The first layer is the embedding layer, with $300$ embedding dimensions, the second layer is Bidirectional CuDNNLSTM, which is a fast LSTM implementation backed by CuDNN, which is a library by NVIDIA CUDA described as a GPU-accelerated library of primitives for deep neural networks~\cite{TensorFlow,NVIDIACUDA}. 
The main advantage of using LSTM is that it preserves the input sequence data. Forward LSTM preserves only the previous data, while using Bidirectional LSTM (BiLSTM) preserves both forward and backward data, so while it understands more information about the context, it requires more computation time, and for this reason we used CuDNNLSTM to reduce the processing time. The third and fourth layers are dense layers, with the first dense layer with linear activation function being chosen by a grid search of relu, sigmoid, linear, and none. Then, there is a second dense layer with a sigmoid activation function. The loss in the compiled model is calculated by Binary Cross Entropy Loss and is optimized by using the Adam optimizer. We tested it over $10$ epochs which were also chosen after grid search over $10$, $20$, and $30$, and we used a $256$ batch size to classify each tweet. We also divided the data sets into $20$\% for testing and $80$\% for training.

\subsection{BERT language model}
\noindent 
The second experiment uses the pre-trained model BERT, which is used to encode the input text. We used the BertForSequenceClassification ~\cite{berttransformersdocumentation} model, in which the last layer is a classification neural-network layer. The task-specific design of BERT is able to represent the input sentence of an array of tokens. For each token, the input representation is composed by summing its corresponding token, segment, and position embeddings. For a classification task, BERT adds a unique token [CLS] at the beginning of the sentence tokens; this token is used as the starting position for the fully connected layer to the last encoder layer, and then a softmax layer to classify the sentence.
 \par 
 There are several versions of BERT which differ according to the language (Chinese, English, Multilingual), the alphabet (Cased, Uncased), and the size of the construction layers (BERT-Base, BERT-Large). We used the BERT-Base model which contains $12$ transformer layers, for each transformer,$12$ self-attention heads and hidden states size is $768$. In comparison, the BERT-Large model contains $24$ transformer layers and $16$ self-attention heads, and the hidden states size is $1024$. The model specifications are: LEARNING\_RATE = $2e-5$, NUM\_TRAIN\_EPOCHS = $3.0$ and BATCH\_SIZE = $32$. We updated BATCH\_SIZE = $16$ for the large model to avoid memory issues, because BERT\-Large needs much more memory than BERT-Base. We ran the code on Google Colab using a GPU for fast execution. 
 \section {Datasets}
 \label{sec:DS}
\noindent 
This section describes the datasets we used in the experiments. We experimented on two datasets: a Stormfront dataset collected from white supremacist extremist content, which is available online (the Stormfront forum was later deleted because of its support for racial hate), and Twitter dataset (Twitter White Supremacy Dataset) to assess the performance on recent white supremacist tweets.
\subsection{Available Dataset (Stormfront Dataset)}
\noindent 
For detecting white supremacists, to the best of our knowledge, there is no dataset available for white supremacy content except the Stormfront dataset~\cite{de2018hate}, which is a dataset collected from the Stormfront white supremacist forum. Random sets of posts have been sampled from several sub-forums and split into sentences, and those sentences have been manually labeled as ‘containing hate speech or not’, ‘skip’, or relation (relation means it needs extra context to annotate), according to certain annotation guidelines. The average Cohen's kappa annotator agreement score for a batch of $1,144$ sentences of the dataset is $61.4$ for three classes (i.e., hate, non-hate, skip) and $62.7$ for four categories (i.e., hate, non-hate, skip and relation) for $1,018$ sentences of the dataset. The Cohen' kappa percentage does not represent the entire published dataset, and is calculated for three or four classes. Their classification experiment was performed on a balanced subset of the dataset which included only hate and non-hate and excluded other classes.
\subsection{Twitter White Supremacy Dataset}	  
\noindent 
The aforementioned dataset was obtained from the Stormfront website, which has been taken offline and no longer available for research purposes. Thus, and to assemble white supremacist posts from different platforms, we collected a dataset from Twitter by randomly selecting subsets of tweets from the white supremacist corpus. The dataset consists of $1,999$ tweets that were annotated by three judges through Amazon Mechanical Turk (AMT). The judges have to be located in North America and have a hit approval greater than $80$\%. The location criterion was chosen to ensure that the reader/annotators fully understood common cultural terminologies, events, figures, and coded words.
\par
The annotation procedure initially consisted of four labels: explicit white supremacy, implicit white supremacy, other hate speech, and neutral. Explicit white supremacy content refers to hate speech/tweets that express either racial or religious hatred towards others or claims of being undermined by other racial or religious groups. For example, “These people do not want solution They only want you dead White man.” Implicit white supremacy content refers to textual information that expresses racial or religious hatred either indirectly or implicitly, e.g., “we own our diversity, leave our country.” Other hate speech is any hateful text other than white supremacist hate text, such as misogyny (i.e., hatred of women), homophobia (i.e., hatred of LGBT people), or sexism (i.e., discrimination based on gender). An example of misogynist hate speech is, “You never gonna find a woman that was down for you like I was there are plenty women as stupid as you in this world dear, I’m sure he’ll be fine.” Neutral text, on the other hand, is any content that expresses positive subjective content (e.g., “Always brother”), factual text (e.g., weather situation), or any other content not intended to promote or encourage hatred. Neutral also includes textual information that is challenging and cannot be annotated as hate speech or non-hate speech due to ambiguous intentions or contexts, e.g., "to be against immigration does not mean to kill people" or "die for them". Also, any factual text that includes hate terms with no hate intent is considered as Neutral (e.g., “Christchurch mosque shooter to be sentenced on August 24”).
\par
The annotators’ agreements for the four labels were very low, with a 0.0706 Cohen’s kappa score. This is because there were large numbers of disagreements between the annotators, especially regarding neutral and implicit white supremacism. The disagreements between the judges were analyzed by counting the number of conflicts between pairs of annotators, and the average numbers of conflict between the annotators were estimated (i.e., conflicts between annotators 1 and 2, annotators 2 and 3, and annotators 1 and 3). We found that the highest number of disagreements involved neutral and implicit white supremacy, with 247 disagreements out of 1,999, by finding the average number of disagreements between the three annotators. For example, two annotators considered a tweet as implicit white supremacist content, while the third annotator considered it as neutral, as shown in Table~\ref{table:t2}. The disagreements between annotators often occurred when the intention of the writer was inexplicit. In the first example, the annotators disagreed on whether the writer had intended to discuss some statistical and factual information about immigration or to promote hatred against immigrants. This ambiguous content can reduce the agreement between annotators, increases the difficulty of detecting harmful content based on the content of the tweet, and requires the entire user profile to give an additional indication about the user’s intent. Annotation is a difficult task because it is highly subjective. Other examples of tweets that were found to be challenging for the annotators can be found in Table~\ref{table:t2}.
\begin{table}[ht]
\caption{Examples of annotators’ disagreements}
\label{table:t2}
\resizebox{\textwidth}{!}{%
\begin{tabular}{|l|l|l|l|l|}
\hline
\textbf{No.} &
  \textbf{Tweet} &
  \textbf{Annotator 1} &
  \textbf{Annotator 2} &
  \textbf{Annotator 3} \\ \hline
1. &
  \begin{tabular}[c]{@{}l@{}}“Carrying capacity is irrelevant, our immigration rate is basically the highest in the\\  OECD 5x the historical NOM rate, 100s of per year is radically changing the\\  demographic composition which is nothing but an insane destructive experiment”\end{tabular} &
  Neutral &
  \begin{tabular}[c]{@{}l@{}}Implicit \\ White \\ Supremacism\end{tabular} &
  \begin{tabular}[c]{@{}l@{}}Implicit\\ White\\ Supremacism\end{tabular} \\ \hline
2. &
  “wonder if any of the African Dictators have her penciled in” &
  \begin{tabular}[c]{@{}l@{}}Implicit \\ White \\ Supremacism\end{tabular} &
  \begin{tabular}[c]{@{}l@{}}Implicit\\ White\\ Supremacism\end{tabular} &
  Neutral \\ \hline
3. &
  \begin{tabular}[c]{@{}l@{}}“This really confuses me not seeing these claims she speaks of \\ but am seeing Christians being murdered”\end{tabular} &
  \begin{tabular}[c]{@{}l@{}}Implicit \\ White\\ Supremacism\end{tabular} &
  Neutral &
  Neutral \\ \hline
\end{tabular}%
}
\end{table}
\par 

To analyze the disagreements, ‘implicit white supremacist’ was removed from the dataset, the dataset size becomes $1,010$ tweets, and the Cohen kappa score becomes $0.1047$, which is much better than the four-label dataset. In this research, we treated the problem as a binary classification problem; thus, we collapse the four labels to binary labels (white supremacy or non-white supremacy). Explicit and implicit white supremacy were collapsed into a ‘white supremacy’ label, and the ‘neutral’ and other hate categories were collapsed into ‘non-white supremacy’ because our goal was to detect white supremacists in particular. We calculated the Cohen’s kappa coefficient for the two labels ($0$ for non-white supremacy, $1$ for white supremacy) for all the annotators, and the agreement score was very low, with a 0.11 Cohen’s kappa score. The disagreement is due to the difficulty in detecting implicit white supremacy hate speech. Schmidt et al.~\cite{schmidt2017survey} recognized from previous studies that the hate speech annotation process is reasonably ambiguous, which results into low agreement scores. To handle the annotators’ disagreements, we used a voting strategy by choosing the most common label among the three annotators, so if at least two annotators agreed on one label, either $0$ or $1$, this label will be used as the final tweet label. 

\subsection{Balanced Combined Dataset}
\noindent 
We created a combined balanced dataset from the datasets used (Twitter and Stormfront) to test the model on the largest possible diversity of the data from different social platforms in order to train the model on the greatest possible white supremacist hate speech. We combined the Stormfront and Twitter datasets by aggregating them into one CSV file, and then balanced them according to the number of class with lower frequency, and randomly selecting other class examples. Table~\ref{table:t3} includes the details of the white supremacist datasets.
\begin{table}[ht]
\caption{Details of white supremacist datasets}
\label{table:t3}
\resizebox{\textwidth}{!}{%
\begin{tabular}{|l|l|l|l|l|}
\hline
\textbf{Dataset} &
  \textbf{Original labels} &
  \textbf{\begin{tabular}[c]{@{}l@{}}\# white\\  supremacist \\ hate\end{tabular}} &
  \textbf{\begin{tabular}[c]{@{}l@{}}\# nonwhite\\  supremacist\\  hate\end{tabular}} &
  \textbf{Total} \\ \hline
\begin{tabular}[c]{@{}l@{}}Stormfront dataset \\ \cite{de2018hate}\end{tabular}   & Hate, non-hate & 1,196 & 9,748 & 10,944 \\ \hline
\begin{tabular}[c]{@{}l@{}}Twitter white\\  supremacist dataset\end{tabular} &
  \begin{tabular}[c]{@{}l@{}}Explicit white supremacist, \\ implicit white supremacist, \\ other hate speech and neutral\end{tabular} &
  1,100 &
  899 &
  1,999 \\ \hline
\begin{tabular}[c]{@{}l@{}}Combined white\\  supremacist balanced\\ dataset\end{tabular} & Hate, non-hate & 2,294 & 2,294 & 4,588  \\ \hline
\end{tabular}%
}
\end{table}

\section {Experiments and Results}
\label{sec:ER}
\noindent We applied two different experiments and evaluated them separately. The first experiment used white supremacist domain-specific embeddings and the Bidirectional LSTM deep model, and the second experiment used the full pre-trained language model(BERT). The experiments were run on Google Colab to use GPU processor for fast execution.
\subsection{Domain-Specific White Supremacy Word Embedding and Deep Model Experiment}
\noindent
This experiment used (WSW2V) embedding as features and the bidirectional LSTM deep model as a classifier. We also used the domain-agnostic embedding models, which were the GloVe pre-trained word-embedding and Google pre-trained embedding models, to test them with the Bidirectional LSTM deep model and compare them against domain-specific (WSW2V) embedding. Table~\ref{table:t4} shows descriptions of the embedding models used.
\begin{table}[ht]
\caption{Details description of embedding models}
\label{table:t4}
\resizebox{\textwidth}{!}{%
\begin{tabular}{|l|l|l|l|}
\hline
\textbf{Methods} &
  \textbf{Dimension} &
  \textbf{Trained on data of size} &
  \textbf{Pretrained on platform} \\ \hline
  GoogleNews-vectors-negative \footnotemark  &
  300 &
  3 billion words &
  Google News \\ \hline
GloVe.6B.300d\footnotemark &
  300 &
  \begin{tabular}[c]{@{}l@{}}6B tokens, \\ 400K vocab,\\ uncased\end{tabular} &
  \begin{tabular}[c]{@{}l@{}}Wikipedia 2014 + \\ English Gigaword Fifth Edition\end{tabular} \\ \hline
GloVe.Twitter.27B.200d\footnotemark[2] &
  200 &
  \begin{tabular}[c]{@{}l@{}}2B tweets, \\ 27B tokens,\\ 1.2M vocab, \\ uncased\end{tabular} &
  Twitter \\ \hline
\begin{tabular}[c]{@{}l@{}}White supremacy Word2Vec \\ (WSW2V)\end{tabular} &
  300 &
  \begin{tabular}[c]{@{}l@{}}1,041,576 tweets, \\ 117083 vocab, \\ uncased\end{tabular} &
  Twitter \\ \hline
\end{tabular}%
}
\end{table}
\footnotetext[1] {\url{https://code.google.com/archive/p/word2vec/}}
\footnotetext[2] {\url{https://nlp.stanford.edu/projects/glove/}}

\par
Table~\ref{table:t5} compares the models’ performance using different embeddings (domain-specific and domain-agnostic), and different classifiers (LR and Bidirectional LSTM deep Model). First, we used WSW2V embedding with Logistic Regression (LR). This LR-WSW2V model performed well on two datasets, but very poorly on the Stormfront forum dataset. Then, we used domain-agnostic word embedding and compared it with domain-specific (WSW2V) embedding using the same classifier (Bidirectional LSTM deep Model). The results show that our pre-trained word embedding (WSW2V) outperformed the other models except for the Twitter white supremacy dataset. The inconsistent performance of the model is due to the class imbalance in the datasets (Stormfront and Twitter), as shown in Table~\ref{table:t3}. At the same time, the word embedding (WSW2V) is biased toward hate speech, as the embedding was extracted from a white supremacist corpus with a majority of white supremacist hate tweets. To overcome this, we created a balanced dataset and experimented with it. We noticed that the Stormfront dataset had very bad performance among the other datasets, and this is because the WSW2V embeddings are extracted from Tweets and not from the forum comments. 
\begin{table}[ht]
\caption{Classification experiment results for domain-specific white supremacy }
\label{table:t5}
\resizebox{\textwidth}{!}{%
\begin{tabular}{|l|l|l|l|l|l|l|}
\hline
\multicolumn{1}{|c|}{\textbf{\begin{tabular}[c]{@{}c@{}}Machine \\ Learning\\ Approach\end{tabular}}} &
  \multicolumn{1}{c|}{\textbf{Word Embedding}} &
  \multicolumn{1}{c|}{\textbf{Dataset}} &
  \multicolumn{1}{c|}{\textbf{Precision}} &
  \multicolumn{1}{c|}{\textbf{Recall}} &
  \multicolumn{1}{c|}{\textbf{F1-score}} &
  \multicolumn{1}{c|}{\textbf{AUC}} \\ \hline
\multirow{3}{*}{\begin{tabular}[c]{@{}l@{}}Logistic Regression \\ (LR)\end{tabular}} &
  \multirow{3}{*}{\begin{tabular}[c]{@{}l@{}}White Supremacy\\ Word2Vec \\ (WSW2V)(300)\end{tabular}} &
  Stormfront dataset &
  0.70871 &
  0.14344 &
  0.23753 &
  0.85424 \\ \cline{3-7} 
 &                                & Twitter white supremacist dataset & 0.6831  & 0.77352 & \textbf{0.72548} & 0.72894 \\ \cline{3-7} 
 &                                & Balanced combined                 & 0.727   & 0.75341 & 0.73996          & 0.8116  \\ \hline
\multirow{12}{*}{\begin{tabular}[c]{@{}l@{}}Bidirectional \\ LSTM\end{tabular}} &
  \multirow{3}{*}{\begin{tabular}[c]{@{}l@{}}GoogleNews-\\ vectors-\\ negative300\end{tabular}} &
  Stormfront dataset &
  0.45161 &
  0.34146 &
  0.38888 &
  0.64448 \\ \cline{3-7} 
 &                                & Twitter white supremacist dataset & 0.61904 & 0.57073 & 0.5939           & 0.60075 \\ \cline{3-7} 
 &                                & Balanced combined                 & 0.72572 & 0.68109 & 0.7027           & 0.72259 \\ \cline{2-7} 
 & \multirow{3}{*}{GloVe.6B.300d} & Stormfront dataset                & 0.53594 & 0.34309 & 0.41836          & 0.65334 \\ \cline{3-7} 
 &                                & Twitter white supremacist dataset & 0.66938 & 0.74545 & 0.70537          & 0.64772 \\ \cline{3-7} 
 &                                & Balanced combined                 & 0.74561 & 0.74074 & 0.74316          & 0.744   \\ \cline{2-7} 
 &
  \multirow{3}{*}{\begin{tabular}[c]{@{}l@{}}GloVe.Twitter.27B\\ .200d\end{tabular}} &
  Stormfront dataset &
  0.38305 &
  0.4728 &
  0.42322 &
  0.68973 \\ \cline{3-7} 
 &                                & Twitter white supremacist dataset & 0.67782 & 0.73636 & \textbf{0.70588} & 0.65429 \\ \cline{3-7} 
 &                                & Balanced combined                 & 0.74242 & 0.74727 & 0.74484          & 0.744   \\ \cline{2-7} 
 &
  \multirow{3}{*}{\begin{tabular}[c]{@{}l@{}}White Supremacy \\ Word2Vec \\  (WSW2V)(300)\end{tabular}} &
  Stormfront dataset &
  0.50892 &
  0.47698 &
  \textbf{0.49244} &
  0.71028 \\ \cline{3-7} 
 &                                & Twitter white supremacy dataset   & 0.67179 & 0.59545 & 0.63132          & 0.61994 \\ \cline{3-7} 
 &                                & Balanced combined                 & 0.75054 & 0.74727 & \textbf{0.7489}  & 0.74945 \\ \hline
\end{tabular}%
}
\end{table}
\par
We evaluated the results of the proposed approach against similar research efforts in the field; however, the only study that had analyzed white supremacy content to detect hate speech is ~\cite{de2018hate} study. The authors released the Stormfront dataset for research use, but they only reported the results for a sample ($2,000$ sentences) of the dataset. Thus, we randomly sampled a balanced subset from that Stormfront dataset. The results show that the Bidirectional LSTM outperformed~\cite{de2018hate} with an accuracy of $0.80$ (only the accuracy is reported in de Gibert et al.'s study). This result shows that our proposed model outperforms their model by $2$ points, given that they used random word embedding for features and LSTM for classification, as shown in Table~\ref{table:t5}.
\begin{table}[ht]
\caption{Domain-specific word embeddings and deep model compared with \cite{de2018hate} evaluation}
\label{table:t6}
\resizebox{\textwidth}{!}{%
\begin{tabular}{|c|c|c|c|c|c|c|c|}
\hline
Machine Learning Approach &
  Word Embedding &
  Dataset &
  \begin{tabular}[c]{@{}c@{}}Dataset \\ size\end{tabular} &
  \begin{tabular}[c]{@{}c@{}}Accuracy\\  (Hate)\end{tabular} &
  \begin{tabular}[c]{@{}c@{}}Accuracy\\  (non- Hate)\end{tabular} &
  Accuracy &
  F1-score \\ \hline
\begin{tabular}[c]{@{}c@{}}LSTM \\ \cite{de2018hate}\end{tabular} &
  \begin{tabular}[c]{@{}c@{}}Random word \\ embedding\end{tabular} &
  \begin{tabular}[c]{@{}c@{}}Balanced subset of\\ de Gibert et al. 's,\\ (2018) dataset\end{tabular} &
  2000 &
  0.76 &
  0.8 &
  0.78 &
  - \\ \hline
\multirow{4}{*}{\begin{tabular}[c]{@{}c@{}}Bidirectional\\  LSTM \\ deep model\end{tabular}} &
  \begin{tabular}[c]{@{}c@{}}GoogleNews-\\ vectors-negative300\end{tabular} &
  \multirow{4}{*}{\begin{tabular}[c]{@{}c@{}}Balanced subset of\\ de Gibert et al.'s,\\ (2018) dataset\end{tabular}} &
  \multirow{4}{*}{2000} &
  0.824 &
  0.6847 &
  0.76 &
  0.7876 \\ \cline{2-2} \cline{5-8} 
 &
  GloVe.6B.300d &
   &
   &
  0.79 &
  0.78 &
  0.785 &
  0.7912 \\ \cline{2-2} \cline{5-8} 
 &
  \begin{tabular}[c]{@{}c@{}}GloVe.Twitter\\ .27B.200d\end{tabular} &
   &
   &
  0.795 &
  0.775 &
  0.785 &
  0.789 \\ \cline{2-2} \cline{5-8} 
 &
  \begin{tabular}[c]{@{}c@{}}White Supremacy \\ Word2Vec  (WSW2V)\\ (300)\end{tabular} &
   &
   &
  0.805 &
  0.8 &
  \textbf{0.8025} &
  0.7925 \\ \hline
\end{tabular}%
}
\end{table}
\subsection{BERT pre-trained language model Experiment}
\noindent 
The second experiment was performed based on the BERT language model, since it has proven to have high performance for many NLP tasks. We used both the BERT Base and Large models. The results of the evaluation is reported in Table~\ref{table:t6}. As shown in the table, BERT provides better performance (F1-score) than our domain-specific model for white supremacist classification. It improved the F1-score by $15$ points on the Stormfront dataset and by about $5$ points on the Twitter and balanced datasets. Also, the Large model did not outperform the Base model for some datasets given that the batch size was reduced. However, BERT-Base provided consistent and high performance for most datasets, and the difference with BERT-Large on the balanced dataset did not exceeds $0.01$ points. This argues that Bert-Base has the best performance among all the models used in this paper. 
\begin{table}[ht]
\caption{BERT sequence classification white supremacist experiment results (Base-Large)}
\label{table:t7}
\resizebox{\textwidth}{!}{%
\begin{tabular}{|l|l|l|l|l|l|}
\hline
\textbf{Methods}           & \textbf{Dataset}   & \textbf{Precision} & \textbf{Recall} & \textbf{f1-score} & \textbf{AUC} \\ \hline
\multirow{3}{*}{BERT Base} & Stormfront dataset & 0.66517            & 0.62869         & \textbf{0.64642}  & 0.79512      \\ \cline{2-6} 
                            & Twitter white supremacist dataset & 0.67843 & 0.8398  & \textbf{0.75054} & 0.70746 \\ \cline{2-6} 
                            & Balanced combined                 & 0.82117 & 0.76034 & 0.78959          & 0.7972  \\ \hline
\multirow{3}{*}{BERT Large} & Stormfront dataset                & 0.63589 & 0.54625 & 0.58767          & 0.75502 \\ \cline{2-6} 
                            & Twitter white supremacist dataset & 0.65    & 0.72558 & 0.68571          & 0.63452 \\ \cline{2-6} 
                            & Balanced combined                 & 0.81573 & 0.7773  & \textbf{0.79605} & 0.79753 \\ \hline
\end{tabular}%
}
\end{table}
\par 
In Table~\ref{table:t8}, we also compare the BERT model’s accuracy with that of ~\cite{de2018hate}. BERT outperformed their accuracy by 8 points using the Large model, and also outperformed the domain-specific model by 6 points (Table 4).

\begin{table}[ht]
\caption{Bert sequence classification compared with ~\cite{de2018hate} results}
\label{table:t8}
\resizebox{\textwidth}{!}{%
\begin{tabular}{|l|l|l|l|l|l|l|l|}
\hline
\textbf{Source} &
  \textbf{Methodology} &
  \textbf{Dataset} &
  \textbf{\begin{tabular}[c]{@{}l@{}}Dataset \\ size\end{tabular}} &
  \textbf{\begin{tabular}[c]{@{}l@{}}Accuracy \\ (Hate)\end{tabular}} &
  \textbf{\begin{tabular}[c]{@{}l@{}}Accuracy \\ (non- Hate)\end{tabular}} &
  \textbf{Accuracy} &
  \textbf{f1-score} \\ \hline
  ~\cite{de2018hate} &
  \begin{tabular}[c]{@{}l@{}}Word embeddings,\\  LSTM\end{tabular} &
  \begin{tabular}[c]{@{}l@{}}Balanced subset of\\ de Gibert et al.'s, \\(2018) dataset\end{tabular} &
  2000 &
  0.76 &
  0.8 &
  0.78 &
  - \\ \hline
\multirow{4}{*}{Our experiments} &
  \multirow{2}{*}{\begin{tabular}[c]{@{}l@{}}BERT model\\  BASE\end{tabular}} &
  \multirow{4}{*}{\begin{tabular}[c]{@{}l@{}}Balanced subset\\ of de Gibert et al.'s, \\(2018) dataset\end{tabular}} &
  \multirow{4}{*}{2,000} &
  \multirow{2}{*}{0.8366} &
  \multirow{2}{*}{0.8172} &
  \multirow{2}{*}{0.827} &
  \multirow{2}{*}{0.8115} \\
 &
   &
   &
   &
   &
   &
   &
   \\ \cline{2-2} \cline{5-8} 
 &
  \multirow{2}{*}{\begin{tabular}[c]{@{}l@{}}BERT model\\  LARGE\end{tabular}} &
   &
   &
  \multirow{2}{*}{0.8373} &
  \multirow{2}{*}{0.8762} &
  \multirow{2}{*}{\textbf{0.86466}} &
  \multirow{2}{*}{0.8662} \\
 &
   &
   &
   &
   &
   &
   &
   \\ \hline
\end{tabular}%
}
\end{table}
\section{Discussion}
\label{sec:DIS}
\noindent The first approach of domain-specific experiments in  (Table~\ref{table:t6}), the results show that domain-specific embedding with Bidirectional LSTM model outperforms the results of ~\cite{de2018hate} who used randomly initialized word embedding with LSTM. Their accuracy was $78$\%, while our accuracy is $80$\%. Although our model exceeds their accuracy, but we expected much higher accuracy than only 2 points, which means that random initialization does not perform very badly. It is important to mention that white supremacist corpus for the pretrained word embedding was about 1 million tweets, increasing the corpus size would provide better performance, but we were limited by Twitter’s policies. This experiment shows that the Bidirectional LSTM based deep model gave good performance for the white supremacy detection, which contradicts~\cite{liu2018neural}, who said that LSTM did not give a good performance because the length of tweets was limited to 180 characters; however, now it is 280 characters.
\par
From the feature perspective comparison, Table~\ref{table:t5} shows how WSW2V performs in comparison with other domain-agnostic models using the same classifier and datasets; the WSW2V outperforms other models on both the Stormfront and Balanced datasets, but GloVe Twitter outperforms WSW2V, and this is because the big size difference of the data trained on, i.e, ($2B$) for  GloVe Twitter and ($~1M$) for WSW2V. From the classifier perspective comparison, the Bidirectional LSTM-based deep model outperforms LR on two datasets (Stormfront and Balanced), but LR outperforms the Bidirectional LSTM-based deep model on the Twitter dataset. 
\par
The second experiment involved using the BERT model on the dataset to assess its performance on the white supremacist hate speech classification task. As shown in Table~\ref{table:t7}, BERT outperforms all the distributional-based embeddings (Google Word2Vec, GloVe and WSW2V) with the Bidirectional LSTM-based deep model in Table~\ref{table:t5}. This means that the BERT model gives a closer meaningful vector of the words due to its training strategy (deeply bidirectional) and the large corpus trained on. The BERT language model combines the advantages of domain-agnostic and domain-specific embeddings in its training strategy, it is petrained on a large corpus and add extra layer for training your specific task. 
\par
Finally, narcissists often use first-person singular pronouns(e.g.\@, "I" and "me") and profane and aggressive language in their social media communications \cite{dewall2011narcissism},  while individuals with an argumentative personality often comment on other people’s posts or frequently post on similar topics to prove their point. White supremacists usually associate themselves with radical groups by either identifying themselves as a member in their profiles or by encouraging or promoting their ideological perspectives. This study focuses on tweets or textual features to detect white supremacy, and not account for profile features. Thus, we only focus on tweet features that help to identify white supremacists’ characteristics. Further account analysis will be included in future work.
	 
\section {Conclusion and Future work}
\label{sec:CON}
\noindent 
From the experiments, we have shown that a combination of word embedding, and deep learning perform well for the problem of white supremacist hate speech. Some of the datasets are imbalanced to simulate real-world data, and others are balanced to assess the model’s performance under an ideal situation. The BERT model has also proved that it provides the state of art for this problem. For future work, the corpus size will be maximized in order to generate more meaningful embeddings, and experiments will be done on multiclass problems instead of binary class problems and by combining Google Word2Vec and domain-specific Word2Vec.

\section {Acknowledgement}
I would like to thank all the researchers who have made their resources available to the research community. 
\bibliography{mybibfile}

\end{document}